\PassOptionsToPackage{unicode}{hyperref}
\PassOptionsToPackage{hyphens}{url}
\PassOptionsToPackage{dvipsnames,svgnames,x11names}{xcolor}
\documentclass[
]{article}
\usepackage{amsmath,amssymb}
\usepackage{lmodern}
\usepackage{iftex}
\ifPDFTeX
  \usepackage[T1]{fontenc}
  \usepackage[utf8]{inputenc}
  \usepackage{textcomp} % provide euro and other symbols
\else % if luatex or xetex
  \usepackage{unicode-math}
  \defaultfontfeatures{Scale=MatchLowercase}
  \defaultfontfeatures[\rmfamily]{Ligatures=TeX,Scale=1}
\fi
\IfFileExists{upquote.sty}{\usepackage{upquote}}{}
\IfFileExists{microtype.sty}{% use microtype if available
  \usepackage[]{microtype}
  \UseMicrotypeSet[protrusion]{basicmath} % disable protrusion for tt fonts
}{}
\makeatletter
\@ifundefined{KOMAClassName}{% if non-KOMA class
  \IfFileExists{parskip.sty}{%
    \usepackage{parskip}
  }{% else
    \setlength{\parindent}{0pt}
    \setlength{\parskip}{6pt plus 2pt minus 1pt}}
}{% if KOMA class
  \KOMAoptions{parskip=half}}
\makeatother
\usepackage{xcolor}
\usepackage{longtable,booktabs,array}
\usepackage{calc} % for calculating minipage widths
\usepackage{etoolbox}
\makeatletter
\patchcmd\longtable{\par}{\if@noskipsec\mbox{}\fi\par}{}{}
\makeatother
\IfFileExists{footnotehyper.sty}{\usepackage{footnotehyper}}{\usepackage{footnote}}
\makesavenoteenv{longtable}
\usepackage{graphicx}
\makeatletter
\def\maxwidth{\ifdim\Gin@nat@width>\linewidth\linewidth\else\Gin@nat@width\fi}
\def\maxheight{\ifdim\Gin@nat@height>\textheight\textheight\else\Gin@nat@height\fi}
\makeatother
\setkeys{Gin}{width=\maxwidth,height=\maxheight,keepaspectratio}
\makeatletter
\def\fps@figure{htbp}
\makeatother
\providecommand{\tightlist}{%
  \setlength{\itemsep}{0pt}\setlength{\parskip}{0pt}}
\newlength{\cslhangindent}
\newlength{\csllabelwidth}
\newlength{\cslentryspacingunit} % times entry-spacing
\newenvironment{CSLReferences}[2] % #1 hanging-ident, #2 entry spacing
 {% don't indent paragraphs
  \setlength{\parindent}{0pt}
  \ifodd #1
  \let\oldpar\par
  \def\par{\hangindent=\cslhangindent\oldpar}
  \fi
  \setlength{\parskip}{#2\cslentryspacingunit}
 }%
 {}
\usepackage{calc}

\ifLuaTeX
\usepackage[bidi=basic]{babel}
\else
\usepackage[american, bidi=default]{babel}
\fi
\def\languageshorthands#1{}
\ifLuaTeX
  \usepackage{selnolig}  % disable illegal ligatures
\fi
\IfFileExists{bookmark.sty}{\usepackage{bookmark}}{\usepackage{hyperref}}
\IfFileExists{xurl.sty}{\usepackage{xurl}}{} % add URL line breaks if available
\hypersetup{
  pdftitle={A C++ Implementation of a Cartesian Impedance Controller for
Robotic Manipulators},
  pdfauthor={Matthias Mayr, Julian M. Salt-Ducaju},
  pdflang={en-US},
  colorlinks=true,
  linkcolor={Maroon},
  filecolor={Maroon},
  citecolor={Blue},
  urlcolor={Blue},
  pdfcreator={LaTeX via pandoc}}

\title{A C++ Implementation of a Cartesian Impedance Controller for
Robotic Manipulators}

\usepackage[affil-it]{authblk}
\usepackage{orcidlink}
\author[1,2%
  ]{Matthias Mayr%
    \,\orcidlink{0000-0002-8198-3154}\,%
    }
\author[2,3%
  ]{Julian M. Salt-Ducaju%
    \,\orcidlink{0000-0001-5256-8245}\,%
    }

\affil[1]{Department of Computer Science, Faculty of Engineering (LTH),
Lund University, Sweden}
\affil[2]{Wallenberg AI, Autonomous Systems and Software Program (WASP),
Sweden}
\affil[3]{Department of Automatic Control, Faculty of Engineering (LTH),
Lund University, Sweden}
\date{19 December 2022}

\begin{document}
\maketitle

\hypertarget{summary}{%
\section{Summary}\label{summary}}

A Cartesian impedance controller is a type of control strategy that is
used in robotics to regulate the motion of robot manipulators. This type
of controller is designed to provide a robot with the ability to
interact with its environment in a stable and compliant manner.
Impedance control increases the safety in contact-rich environments by
establishing a mass-spring-damper relationship between the external
forces acting on the robot and the variation from its reference defined
by a set of coordinates that describe the motion of a robot. As a
consequence, the controlled robot behaves in a compliant way with
respect to its external forces, which has the added benefit of allowing
a human operator to interact with the robot, such as manually guiding
it.

In this package, we provide a C++ implementation of a controller that
allows collaborative robots:

\begin{enumerate}
\def\labelenumi{\arabic{enumi}.}
\tightlist
\item
  To achieve compliance in its Cartesian task-frame coordinates.
\item
  To allow for joint compliance in the nullspace of its task-frame
  coordinates.
\item
  To be able to apply desired forces and torques to the environment of
  the robot, \emph{e.g.}, for direct force control.
\end{enumerate}

This package can be used in any torque-controlled robotic manipulator.
Its implementation in Robot Operating System (ROS) integrates it into
\texttt{ros\_control} (\protect\hyperlink{ref-ros_control}{Chitta et
al., 2017}) and can automatically utilize the URDF description of the
robot's geometry.

\hypertarget{statement-of-need}{%
\section{Statement of Need}\label{statement-of-need}}

Modern robotics is moving more and more past the traditional robot
systems that have hard-coded paths and stiff manipulators. Many
use-cases require the robots to work in semi-structured environments.
These environments impose uncertainties that could cause collisions.
Furthermore, many advanced assembly, manufacturing and household
scenarios such as insertions or wiping motions require the robot to
excert a controlled force on the environment. Finally, the robot
workspace is becoming increasingly shared with human workers in order to
leverage both agents and allow them to complement each other.

An implementation of compliant control for robotic manipulators is an
attractive solution for robots in contact-rich environments. To cover
the wide variety of tasks and scenarios, we think that it needs to
fulfill the following criteria:

\begin{enumerate}
\def\labelenumi{\arabic{enumi}.}
\tightlist
\item
  Dynamically adapt the end-effector reference point.
\item
  Dynamically adjust the robot's impedance (\emph{i.e.}, its ability to
  resist or comply with external forces).
\item
  Apply commanded forces and torques (\emph{i.e.}, a wrench) with the
  end-effector of the robot.
\item
  Command a joint configuration and apply it in the nullspace of the
  Cartesian robotic task.
\item
  Execute joint-space trajectories.
\end{enumerate}

A complete implementation with respect to items 1-5 above of compliance
for torque-commanded robotic manipulators is not available, and the
existing solutions \texttt{franka\_ros}
(\protect\hyperlink{ref-franka_ros}{Franka Emika, 2017a}) and
\texttt{libfranka} (\protect\hyperlink{ref-libfranka}{Franka Emika,
2017b}) as well as the \texttt{KUKA} FRI Cartesian impedance controller
can only be used for a single type of robotic manipulator:

\begin{longtable}[]{@{}
  >{\raggedright\arraybackslash}p{(\columnwidth - 8\tabcolsep) * \real{0.3256}}
  >{\centering\arraybackslash}p{(\columnwidth - 8\tabcolsep) * \real{0.2442}}
  >{\centering\arraybackslash}p{(\columnwidth - 8\tabcolsep) * \real{0.1395}}
  >{\centering\arraybackslash}p{(\columnwidth - 8\tabcolsep) * \real{0.1279}}
  >{\centering\arraybackslash}p{(\columnwidth - 8\tabcolsep) * \real{0.1628}}@{}}
\toprule()
\begin{minipage}[b]{\linewidth}\raggedright
\end{minipage} & \begin{minipage}[b]{\linewidth}\centering
KUKA FRI controller
\end{minipage} & \begin{minipage}[b]{\linewidth}\centering
franka\_ros
\end{minipage} & \begin{minipage}[b]{\linewidth}\centering
libfranka
\end{minipage} & \begin{minipage}[b]{\linewidth}\centering
\textbf{This package}
\end{minipage} \\
\midrule()
\endhead
Reference Pose Update &
(x)\footnote{\label{fri}An FRI connection can send either joint position updates or wrench updates}
& x & & \textbf{x} \\
Cartesian Stiffness Update & & x & & \textbf{x} \\
Cartesian Wrench Update & (x)\textsuperscript{\ref{fri}} & & &
\textbf{x} \\
Nullspace Control & ? & x & & \textbf{x} \\
Kinesthetic Teaching &
(x)\footnote{Reaching a joint limit triggers a safety stop}\textsuperscript{,}\footnote{\label{stiffness}Can be implemented by setting the Cartesian stiffness to zero}
& x & (x)\textsuperscript{\ref{stiffness}} & \textbf{x} \\
Trajectory Execution & & & & \textbf{x} \\
Multi-Robot Support & & & & \textbf{x} \\
\bottomrule()
\end{longtable}

This implementation offers a base library that can easily be integrated
into other software and also implements a \texttt{ros\_control}
controller on top of the base library for the popular ROS middleware.
The base library can be used with simulation software such as DART
(\protect\hyperlink{ref-Lee2018}{Lee et al., 2018}). It is utilized in
several research papers such as Mayr, Ahmad, et al.
(\protect\hyperlink{ref-mayr22skireil}{2022}), Mayr, Hvarfner, et al.
(\protect\hyperlink{ref-mayr22priors}{2022}) and Ahmad et al.
(\protect\hyperlink{ref-ahmad2022generalizing}{2022}) that explore
reinforcement learning as a strategy to accomplish contact-rich
industrial robot tasks.

The Robot Operating System (ROS) is an open-source middleware that is
widely used in the robotics community for the development of robotic
software systems (\protect\hyperlink{ref-quigley:2009}{Quigley et al.,
2009}). Within ROS, an implementation of compliant control is available
for position-commanded and velocity-commanded robotic manipulators with
the \texttt{cartesian\_controllers} package
(\protect\hyperlink{ref-FDCC}{Scherzinger et al., 2017}). However, if a
robotic manipulator supports direct control of the joint torques,
\emph{e.g.}, the \texttt{KUKA\ LBR\ iiwa} or the
\texttt{Franka\ Emika\ Robot\ (Panda)}, torque-commanded Cartesian
impedance control is often the preferred control strategy, since a
stable compliant behavior might not be achieved for position-commanded
and velocity-commanded robotic manipulators
(\protect\hyperlink{ref-lawrence:1988}{Lawrence, 1988}).

\hypertarget{control-implementation}{%
\section{Control Implementation}\label{control-implementation}}

The gravity-compensated rigid-body dynamics of the controlled robot can
be described, in the joint space of the robot \(q\in \mathbb{R}^{n}\),
as (\protect\hyperlink{ref-springer:2016}{Siciliano \& Khatib, 2016}):
\begin{equation}\label{eq:rigbod_q}
    M(q)\ddot{q} + C(q,\dot{q})\dot{q} = \tau_{\mathrm{c}} + \tau^{\mathrm{ext}}
\end{equation} where \(M(q)\in \mathbb{R}^{n\times n}\) is the
generalized inertia matrix, \(C(q,\dot{q})\in \mathbb{R}^{n\times n}\)
captures the effects of Coriolis and centripetal forces,
\(\tau_{\mathrm{c}}\in \mathbb{R}^{n}\) represents the input torques,
and \(\tau^{\mathrm{ext}}\in \mathbb{R}^{n}\) represents the external
torques, with \(n\) being the number of joints of the robot. Since the
proposed controller was evaluated using robots that are automatically
gravity-compensated (\texttt{KUKA\ LBR\ iiwa} and
\texttt{Franka\ Emika\ Robot\ (Panda)}), the gravity-induced torques
have not been included in (\autoref{eq:rigbod_q}). However, the proposed
controller can be used in robots that are not automatically
gravity-compensated by adding a gravity-compensation term to the
commanded torque signal, \(\tau_{\mathrm{c}}\).

Moreover, the torque signal commanded by the proposed controller to the
robot, \(\tau_{\mathrm{c}}\) in (\autoref{eq:rigbod_q}), is composed by
the superposition of three joint-torque signals:
\begin{equation}\label{eq:tau_c}
    \tau_{\mathrm{c}} = \tau_{\mathrm{c}}^\mathrm{ca} + \tau_{\mathrm{c}}^\mathrm{ns} + \tau_{\mathrm{c}}^\mathrm{ext}
\end{equation} where

\begin{itemize}
\item
  \(\tau_{\mathrm{c}}^\mathrm{ca}\) is the torque commanded to achieve a
  Cartesian impedance behavior
  (\protect\hyperlink{ref-hogan:1985}{Hogan, 1985}) with respect to a
  Cartesian pose reference in the \(m\)-dimensional task space,
  \(\xi^{\mathrm{D}}\in\mathbb{R}^{m}\), in the frame of the
  end-effector of the robot: \begin{equation}\label{eq:tau_sup}
       \tau_{\mathrm{c}}^\mathrm{ca} = J^{\mathrm{T}}(q)\left[-K^\mathrm{ca}\Delta \xi-D^\mathrm{ca}J(q) \dot{q}\right]
   \end{equation} with \(J(q)\in \mathbb{R}^{m \times n}\) being the
  Jacobian relative to the end-effector (task) frame of the robot, and
  \(K^\mathrm{ca}\in \mathbb{R}^{m \times m}\) and
  \(D^\mathrm{ca}\in \mathbb{R}^{m \times m}\) being the virtual
  Cartesian stiffness and damping matrices, respectively. Also, the
  Cartesian pose error, \(\Delta \xi\) in (\autoref{eq:tau_sup}) is
  defined as
  \(\Delta \xi_{\mathrm{tr}} = \xi_{\mathrm{tr}}-\xi_{\mathrm{tr}}^{\mathrm{D}}\)
  for the translational degrees of freedom of the Cartesian pose and as
  \mbox{$\Delta \xi_{\mathrm{ro}} = \xi_{\mathrm{ro}}\left(\xi_{\mathrm{ro}}^{\mathrm{D}}\right)^{-1}$}
  for the rotational degrees of freedom.
\item
  \(\tau_{\mathrm{c}}^\mathrm{ns}\) is the torque commanded to achieve a
  joint impedance behavior with respect to a desired configuration and
  projected in the null-space of the robot's Jacobian, to not affect the
  Cartesian motion of the robot's end-effector
  (\protect\hyperlink{ref-ott:2008}{Ott, 2008}):
  \begin{equation}\label{eq:tau_ns}
        \tau_{\mathrm{c}}^\mathrm{ns} = \left(I_n-J^{\mathrm{T}}(q)(J^{\mathrm{T}}(q))^\mathrm{\dagger}\right)\tau_0
    \end{equation} with the superscript \(^\mathrm{\dagger}\) denoting
  the Moore-Penrose pseudoinverse matrix
  (\protect\hyperlink{ref-khatib:1995}{Khatib,
  1995})\footnote{The Moore-Penrose pseudoinverse is computationally cheap and allows a null-space projection disregarding the dynamics of the robot. However, not using the dynamics of the robot to fomulate a pseudoinverse matrix for null-space projection may cause that a non-zero arbitrary torque, $\tau_0$ in (\autoref{eq:tau_ns}), generates interfering forces in the Cartesian space if the joint of the robot are not in a static equilibrium ($\dot{q} = \ddot{q} = 0$).}
  given by \mbox{$J^\dagger = (J^\mathrm{T}J)^{-1}J^\mathrm{T}$}
  (\protect\hyperlink{ref-ben:2003}{Ben-Israel \& Greville, 2003}), and
  \(\tau_0\) being an arbitrary joint torque formulated to achieve joint
  compliance, \begin{equation}\label{eq:tau_0}
        \tau_0 = -K^\mathrm{ns}(q-q^{\mathrm{D}}) - D^\mathrm{ns} \dot{q}
    \end{equation} where \(K^\mathrm{ns}\in \mathbb{R}^{n \times n}\)
  and \(D^\mathrm{ns}\in \mathbb{R}^{n \times n}\) are the virtual joint
  stiffness and damping matrices, respectively.
\item
  \(\tau_{\mathrm{c}}^\mathrm{ext}\) is the torque commanded to achieve
  the desired external force command in the frame of the end-effector of
  the robot, \(F_{\mathrm{c}}^\mathrm{ext}\):
  \begin{equation}\label{eq:tau_ext}
        \tau_{\mathrm{c}}^\mathrm{ext} = J^{\mathrm{T}}(q)F_{\mathrm{c}}^\mathrm{ext}
    \end{equation}
\end{itemize}

\hypertarget{safety-measures}{%
\subsection{Safety Measures}\label{safety-measures}}

\begin{figure}[!ht]
\centering
\includegraphics[width=0.6\textwidth,height=\textheight]{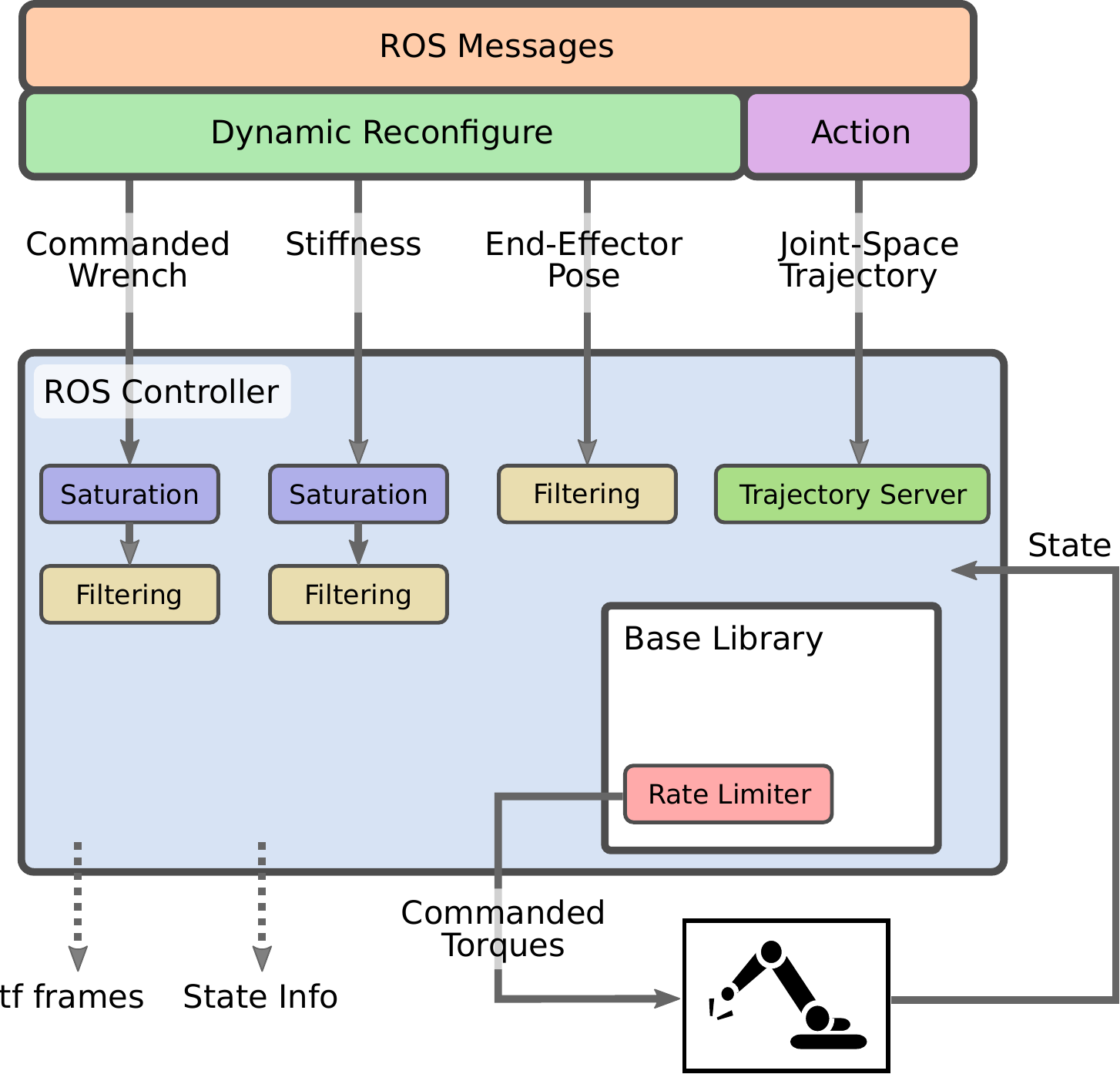}
\caption{Block diagram of the controller. \label{fig:flowchart}}
\end{figure}

As described in \autoref{fig:flowchart}, there are several safety
measures that have been implemented in the controller to achieve a
smooth behavior of the robot:

\hypertarget{filtering}{%
\subsubsection{\texorpdfstring{Filtering
\label{filt}}{Filtering }}\label{filtering}}

The proposed controller allows the online modification of relevant
variables: \(\xi^{\mathrm{D}}\), \(K^\mathrm{ca}\) and \(D^\mathrm{ca}\)
in (\autoref{eq:tau_sup}), \(K^\mathrm{ns}\) and \(D^\mathrm{ns}\) in
(\autoref{eq:tau_0}), and \(F_{\mathrm{c}}^\mathrm{ext}\) in
(\autoref{eq:tau_ext}). However, for a smoother behavior of the
controller, the values of these variables are low-pass filtered. The
update law at each time-step \(k\) is: \begin{equation}
    \alpha_{k+1} = (1-a)\alpha_k + a \alpha^\mathrm{D}
\end{equation} where \(\alpha^\mathrm{D}\) is the desired new variable
value and \(a\in(0,1]\) is defined in such a way that a user-defined
percentage of the difference between the desired value
\(\alpha^\mathrm{D}\) and the variable value at the time of the online
modification instruction, \(\alpha_0\), is applied after a user-defined
amount of time.

\hypertarget{saturation}{%
\subsubsection{Saturation}\label{saturation}}

To increase safety in the controller, some of the filtered variables
(the stiffness and damping factors \(K^\mathrm{ca}\), \(D^\mathrm{ca}\),
\(K^\mathrm{ns}\) and \(D^\mathrm{ns}\), and the desired external force
command \(F_{\mathrm{c}}^\mathrm{ext}\)) can be saturated between
user-defined maximum and minimum limits, \emph{i.e.}, for an example
variable \(\alpha\): \begin{equation}
    \alpha_\mathrm{min} \leq \alpha \leq \alpha_\mathrm{max} 
\end{equation}

\hypertarget{rate-limiter}{%
\subsubsection{Rate Limiter}\label{rate-limiter}}

The rate of the commanded torque, \(\tau_\mathrm{c}\) in
(\autoref{eq:tau_c}), can be limited. For two consecutive commands at
times \(k\) and \(k+1\): \begin{equation}
    \Delta \tau_\mathrm{max} \geq \|\tau_{\mathrm{c},k+1} - \tau_{\mathrm{c},k}\|
\end{equation}

\hypertarget{block-diagram}{%
\section{Block Diagram}\label{block-diagram}}

\hypertarget{acknowledgements}{%
\section{Acknowledgements}\label{acknowledgements}}

We thank Björn Olofsson and Anders Robertsson for the discussions and
feedback. Furthermore, we thank Konstantinos Chatzilygeroudis for the
permission to use the \texttt{RBDyn} wrapper code.

This work was partially supported by the Wallenberg AI, Autonomous
Systems and Software Program (WASP) funded by Knut and Alice Wallenberg
Foundation. The authors are members of the ELLIIT Strategic Research
Area at Lund University.

\hypertarget{references}{%
\section*{References}\label{references}}
\addcontentsline{toc}{section}{References}

\hypertarget{refs}{}
\begin{CSLReferences}{1}{0}
\leavevmode\vadjust pre{\hypertarget{ref-ahmad2022generalizing}{}}%
Ahmad, F., Mayr, M., Topp, E. A., Malec, J., \& Krueger, V. (2022).
Generalizing behavior trees and motion-generator (BTMG) policy
representation for robotic tasks over scenario parameters. \emph{2022
IJCAI Planning and Reinforcement Learning Workshop}.

\leavevmode\vadjust pre{\hypertarget{ref-ben:2003}{}}%
Ben-Israel, A., \& Greville, T. N. (2003). \emph{Generalized inverses:
Theory and applications}. Springer-Verlag, New York, USA.
\url{https://doi.org/10.1007/b97366}

\leavevmode\vadjust pre{\hypertarget{ref-ros_control}{}}%
Chitta, S., Marder-Eppstein, E., Meeussen, W., Pradeep, V., Rodríguez
Tsouroukdissian, A., Bohren, J., Coleman, D., Magyar, B., Raiola, G.,
Lüdtke, M., \& Fernández Perdomo, E. (2017). Ros\_control: A generic and
simple control framework for ROS. \emph{The Journal of Open Source
Software}. \url{https://doi.org/10.21105/joss.00456}

\leavevmode\vadjust pre{\hypertarget{ref-franka_ros}{}}%
Franka Emika. (2017a). \emph{Franka {E}mika {P}anda -- franka\_ros
{D}ocumentation}.
\url{https://frankaemika.github.io/docs/franka_ros.html}.

\leavevmode\vadjust pre{\hypertarget{ref-libfranka}{}}%
Franka Emika. (2017b). \emph{Franka {E}mika {P}anda -- libfranka
{D}ocumentation}.
\url{https://frankaemika.github.io/docs/libfranka.html}.

\leavevmode\vadjust pre{\hypertarget{ref-hogan:1985}{}}%
Hogan, N. (1985). Impedance control: An approach to manipulation: Parts
{I--III}. \emph{J. Dynamic Syst., Measurement, and Control},
\emph{107}(1), 1--24. \url{https://doi.org/10.1115/1.3140702}

\leavevmode\vadjust pre{\hypertarget{ref-khatib:1995}{}}%
Khatib, O. (1995). Inertial properties in robotic manipulation: An
object-level framework. \emph{The International Journal of Robotics
Research}, \emph{14}(1), 19--36.
\url{https://doi.org/10.1177/027836499501400103}

\leavevmode\vadjust pre{\hypertarget{ref-lawrence:1988}{}}%
Lawrence, D. A. (1988). Impedance control stability properties in common
implementations. \emph{IEEE International Conference on Robotics and
Automation (ICRA)}, 1185--1190.
\url{https://doi.org/10.1109/ROBOT.1988.12222}

\leavevmode\vadjust pre{\hypertarget{ref-Lee2018}{}}%
Lee, J., Grey, M. X., Ha, S., Kunz, T., Jain, S., Ye, Y., Srinivasa, S.
S., Stilman, M., \& Liu, C. K. (2018). {DART}: Dynamic animation and
robotics toolkit. \emph{The Journal of Open Source Software},
\emph{3}(22), 500. \url{https://doi.org/10.21105/joss.00500}

\leavevmode\vadjust pre{\hypertarget{ref-mayr22skireil}{}}%
Mayr, M., Ahmad, F., Chatzilygeroudis, K., Nardi, L., \& Krueger, V.
(2022). \emph{Skill-based multi-objective reinforcement learning of
industrial robot tasks with planning and knowledge integration}. arXiv.
\url{https://doi.org/10.48550/ARXIV.2203.10033}

\leavevmode\vadjust pre{\hypertarget{ref-mayr22priors}{}}%
Mayr, M., Hvarfner, C., Chatzilygeroudis, K., Nardi, L., \& Krueger, V.
(2022). Learning skill-based industrial robot tasks with user priors.
\emph{2022 IEEE 18th International Conference on Automation Science and
Engineering (CASE)}, 1485--1492.
\url{https://doi.org/10.1109/CASE49997.2022.9926713}

\leavevmode\vadjust pre{\hypertarget{ref-ott:2008}{}}%
Ott, C. (2008). \emph{Cartesian impedance control of redundant and
flexible-joint robots}. Springer, Berlin, Germany.

\leavevmode\vadjust pre{\hypertarget{ref-quigley:2009}{}}%
Quigley, M., Conley, K., Gerkey, B., Faust, J., Foote, T., Leibs, J.,
Wheeler, R., Ng, A. Y., \& others. (2009). ROS: An open-source robot
operating system. \emph{IEEE International Conference on Robotics and
Automation (ICRA): Workshop on Open Source Software}.

\leavevmode\vadjust pre{\hypertarget{ref-FDCC}{}}%
Scherzinger, S., Roennau, A., \& Dillmann, R. (2017). Forward dynamics
compliance control (FDCC): A new approach to cartesian compliance for
robotic manipulators. \emph{IEEE/RSJ International Conference on
Intelligent Robots and Systems (IROS)}, 4568--4575.
\url{https://doi.org/10.1109/IROS.2017.8206325}

\leavevmode\vadjust pre{\hypertarget{ref-springer:2016}{}}%
Siciliano, B., \& Khatib, O. (2016). \emph{Springer handbook of
robotics}. Springer, Berlin, Germany.
\url{https://doi.org/10.1007/978-3-319-32552-1}

\end{CSLReferences}

\end{document}